\documentclass{article}
\usepackage{spconf,amsmath,graphicx}
\usepackage{multicol}
\usepackage{multirow}
\usepackage{fancyhdr}
\usepackage{todonotes}
\usepackage{tabularx}
\usepackage{booktabs}
\usepackage[acronym]{glossaries}
\usepackage{caption}
\newacronym{ASR}{ASR}{automatic speech recognition}
\newacronym{E2E}{E2E}{end-to-end}
\newacronym{CTC}{CTC}{connectionist temporal classification}
\newacronym{LAS}{LAS}{listen-attend-spell}
\newacronym{RNN}{RNN}{recurrent neural network}
\newacronym{RNNs}{RNNs}{recurrent neural networks}
\newacronym{LSTM}{LSTM}{long short-term memory}
\newacronym{FFN}{FFN}{feed forward neural}
\newacronym{MHSA}{MHSA}{Multi-Head Self-Attention}
\newacronym{MHA}{MHA}{Multi-Head Attention}
\newacronym{MHCA}{MHCA}{Multi-Head Cross Attention}
\newacronym{RNN-T}{RNN-T}{Recurrent Neural Network Transducer}
\newacronym{WER}{WER}{word error rate}
\newacronym{WERR}{WERR}{word error rate reduction}
\newacronym{rWERR}{rWERR}{relative word error rate reduction}
\newacronym{CP}{CP}{Copying}
\newacronym{CA}{CA}{Cross-Attention}
\newacronym{CW}{CW}{Context Window}
\newacronym{FC}{FC}{Feature Concatenation}
\newacronym{SPM}{SPM}{sentence piece model}
\newacronym{BERT}{BERT}{Bidirectional Encoder Representations from Transformers}
\newacronym{LN}{LN}{Layer Normalization}

\title{PromptFormer: Prompted conformer transducer for ASR}

\name{
\begin{tabular}{@{}c@{}}
Sergio Duarte-Torres$^1$, Arunasish Sen$^2$, Aman Rana$^1$ \\
Lukas Drude$^1$, Alejandro Gomez-Alanis$^1$, Andreas Schwarz$^1$, Leif Rädel$^1$, Volker Leutnant$^1$
\end{tabular}}

\address{$^1$Amazon AGI, Aachen, Germany; $^2$Amazon AGI, Cambridge, UK}

\begin{document}
\maketitle
\begin{abstract}
Context cues carry information which can improve multi-turn interactions in \gls{ASR} systems.~In this paper, we introduce a novel mechanism inspired by hyper-prompting to fuse textual context with acoustic representations in the attention mechanism. Results on a test set with multi-turn interactions show that our method achieves 5.9\% \gls{rWERR} over a strong baseline. We show that our method does not degrade in the absence of context and leads to improvements even if the model is trained without context. We further show that leveraging a pre-trained sentence-piece model for context embedding generation can outperform an external BERT model.
\end{abstract}
\begin{keywords}
Conformer transducer, text context modeling, prompting, Automatic Speech Recognition
\end{keywords}
\section{Introduction}
\label{sec:intro}
Cross-utterance context has been shown to be beneficial for improving the accuracy of long-form \cite{context_google_2023} and dialog-oriented \gls{ASR} systems \cite{wei2022improving,10095055}.~This context can refer to previous \gls{ASR} recognitions, system responses, dialog acts \cite{10095055, chang2023dialog,sunder} or other information up to the current point of the speech session \cite{wei2021attentive,Tran_Wei_Ruan_McGowan_Susanj_Strimel_2022}.%
~In this work, we integrate context into the attention mechanism of the encoder of conformer-transducer \gls{ASR} models.~Our goal is to improve \gls{ASR} tasks involving multi-turn interactions. A typical multi-turn interaction example can arise when an user follows up a query like \textit{who is the singer of \textbf{x}?} with \textit{when was \textbf{x} released?} or \textit{can you play a song from \textbf{x}?}.
We propose to account for such textual context like \textit{\textbf{x}} across turns by concatenating textual and acoustic representations in the keys and values of the \gls{MHA} mechanism of the conformer encoder.~This approach draws inspiration from hyper-prompting \cite{he2022hyperprompt}. 
Our approach differs from traditional multi-headed cross-attention methods \cite{jain2020contextual} in that it does not require separate kernels to create textual query, key and values projections. In our method, acoustic and textual kernels are shared, which leads to better fused multimodal representations, as demonstrated by our experiments.
Alternatively, \cite{wei2021attentive} concatenates context representations in the feature dimension of the input.~This method is prohibitive when using a large number of context embeddings and under-performs attention-based methods. Context has also been integrated by expanding the architecture of \gls{E2E} \gls{ASR} systems \cite{jain2020contextual,sathyendra2022contextual,Procter,wei2021attentive,hori,masumura,context_aware_tranformer}.~\cite{wei2021attentive} includes context encoders and a context combiner module, which are trained with a combined loss to predict intents and transcriptions.~\cite{jain2020contextual} adds an embedding extractor to represent sentence pieces of context tokens and an attention module to attend on each token.~\cite{pundak2018deep} introduces a dedicated biasing and attending mechanism module to the \gls{LAS} architecture to bias context words.~\cite{hori} proposes an expanded Transformer ASR with a conformer encoder that uses multiple consecutive utterances as context for monologue and dialogues. Similarly, \cite{masumura} proposes a token and sentence hierarchical transformer text encoder, which is trained by distilling from a pre-trained large context language model.~Although these extensions have proven effective to leverage contextual information, they come at the expense of an increase in model parameters.~They also add complexity to the training regime, leading to increases in latency and cost. In ~\cite{context_google_2023}, it is proposed to fuse context, text and acoustic representations in the joint network using context representations obtained with a \gls{BERT} model. These are summarized with self-attentive pooling.~In this work, we compare context representations obtained with a pretrained \gls{BERT} model against an encoder derived from the embedding layer of the prediction network.~We show that the latter outperforms the representations from the former, with the added advantage of requiring less OPS.~Our method also deviates from \cite{context_google_2023} in that it does not require to train all model parameters to obtain significant improvements.

In section 2, we describe our conformer model architecture as well as the methods we consider for context generation and context consumption. In section 3, we describe our experimental setup, and discuss experimental results in section 4.

\section{METHOD}
\label{sec:method}
\begin{figure}[t]
\begin{minipage}[b]{1.0\linewidth}
  \centering
  \centerline{\includegraphics[width=8.5cm]{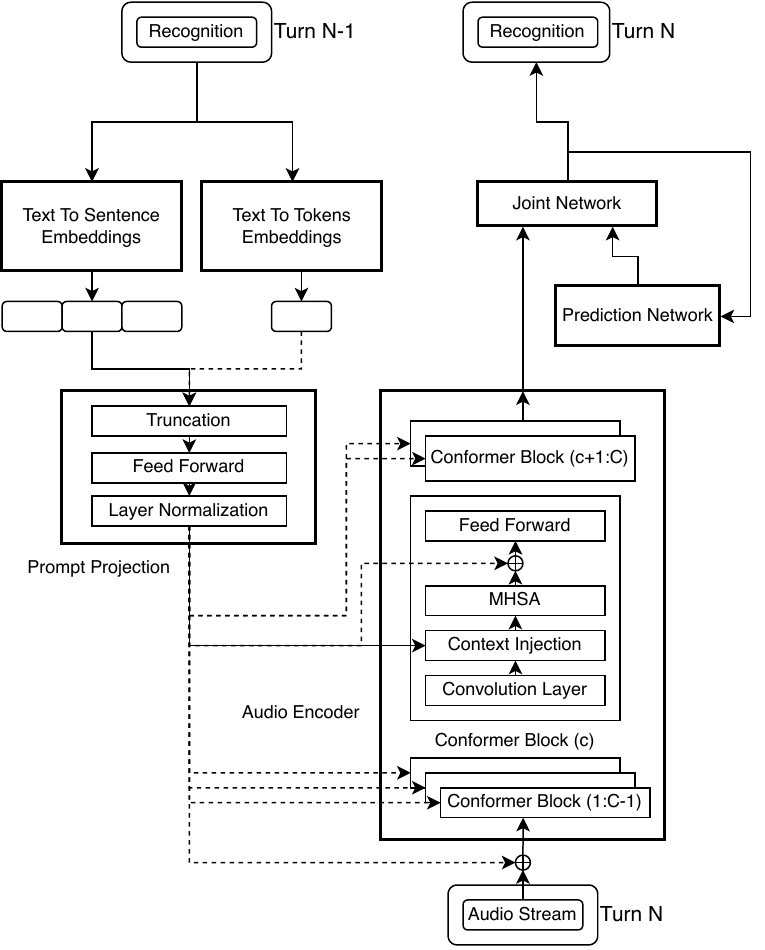}}
\end{minipage}
\caption{\textit{Proposed system for generating and consuming textual context prompts. Dotted lines represent the various ablative configurations described in our result section.}}
\label{fig:system}
\end{figure}
\subsection{Streaming Conformer Transducer}
In this work, the \gls{RNN-T} \cite{rnnt} model structure is used for \gls{ASR}. The neural transducer consists of three modules: (i) audio encoder, (ii) prediction network, and (iii) joint network. Our audio encoder is based on the conformer architecture \cite{conformer}. To use the audio encoder in a streaming setup, we use a causal masking with temporally shifted sliding window masks for each frame of audio.

\subsection{Context Generation}

We use \gls{ASR} recognitions from the previous turns $\mathbf{text}_{rec}^{1:N}$ (which we denote as $\mathbf{text}$ for simplicity, as we use $\mathbf{\text{N}=1}$ for all our experiments) as context source.~We use two approaches to convert the  $\mathbf{text}$ into a token sequence $\mathbf{T}_{1:S}$ of length $S$: by (a) using an external pre-trained \gls{BERT} model, and (b) using our internal pre-trained \gls{SPM}.~For the former, we use a frozen BERT-Tiny model \cite{DBLP:journals/corr/abs-1810-04805} having 2 transformer layers with 128 head size (4.4M parameters) to generate either sentence or token level embeddings.~For the sentence level embeddings, we obtain the CLS token output representation, denoted as $\mathbf{Enc}_{bert}^{CLS}$. For the token level embeddings, $\mathbf{Enc}_{bert}$, we use the corresponding outputs of individual subword units and discard the CLS token output. For the internal pre-trained \gls{SPM} approach, we utilize a learnable embedding layer but initialize the embedding in two different methods: (1) random initialization $\mathbf{Emb}_{ri}$, and (2) initialization from the pre-trained \gls{RNN-T} prediction network embedding layer $\mathbf{Emb}_{pn}$.

The encoded token sequence $\mathbf{T}_{1:S}$ is projected into the encoder's hidden dimension via a series of dense projections $\mathbf{W}_{p}$ and $\mathbf{tanh}$ activations, which has shown beneficial in prompt fine-tuning in NLP tasks \cite{Lliu2022ptuning}.~Finally, we apply a \gls{LN} to obtain a prompt embedding $\mathbf{P}_{1:S}$. For summarization techniques like $\mathbf{Enc}_{bert}^{CLS}$, the sequence length is 1. For other cases, we perform a truncation of the token sequence $\mathbf{P}_{1:S}$ to obtain $\mathbf{P}_{1:TW}$ with fixed token window size $\mathbf{\text{TW}=\mathbf{30}}$ to cater towards runtime latency requirements of our streaming setup. Our final context representation is therefore given by
\begin{equation}
\begin{aligned}
\mathbf{P}_{1:S} &= \mathbf{tanh}_{pn}( ...\mathbf{tanh}_{p1}(\mathbf{T}_{1:S}\mathbf{W}_{p1}) ... \mathbf{W}_{pn}), \\
\mathbf{P}_{1:TW} &= \mathbf{Trunc}_{1:TW}(\text{LN}(\mathbf{P}_{1:S}))
\end{aligned}
\end{equation}
The effective number of trainable weights introduced in both setups (a) and (b) is $\|\mathbf{W}_{p}\|$. The overall number of ops is significantly larger in-case of (a) due to the extra \gls{BERT} encoder.
\subsection{Context Consumption}
Our first context consumption baseline is based on simple concatenation approach, where the context embedding is concatenated to the features of each acoustic frame along the feature dimension, as described in Equation \ref{eq:naive_concat}. This is only applicable when we use a summary embedding like $\mathbf{Enc}_{bert}^{CLS}$.
\begin{equation} \label{eq:naive_concat}
\hat{\mathbf{A}}_{input} = (\mathbf{P}_{1:1} \oplus \mathbf{A}_{input}) \\
\end{equation}
Our second context consumption baseline is inspired from \cite{jain2020contextual} and \cite{pundak2018deep}.
In this case, masked cross-attention is used as a separate biasing module. We use the output of  \gls{MHSA} module at conformer block \textit{c}, denoted as $\mathbf{MHSA}_{output}^{c}$, and $\mathbf{P}_{1:TW}$ to produce the queries, keys and values using separate \gls{CA} kernels $\mathbf{W}_{qca}$, $\mathbf{W}_{kca}$, $\mathbf{W}_{vca}$ respectively. These are used in a \gls{MHA} module to produce $\mathbf{MHCA}_{output}^{c}$ for block \textit{c}. This is then combined back with $\mathbf{MHSA}_{output}^{c}$ and passed to the corresponding feed forward layer. The entire operation is described in Equation \ref{eq:mcha}. 
\begin{equation}\label{eq:mcha}
\begin{aligned}
\mathbf{kca}^{c} &= (\mathbf{P}_{1:TW}) \mathbf{W}_{kca}, \\
\mathbf{vca}^{c} &= (\mathbf{P}_{1:TW}) \mathbf{W}_{vca}, \\
\mathbf{qca}^{c} &=  (\mathbf{MHSA}_{output}^{c}) \mathbf{W}_{qca}, \\
\mathbf{MHCA}_{output}^{c} &= \textbf{MHA}(\mathbf{qca}^{c}, \mathbf{kca}^{c},\mathbf{vca}^{c}), \\
\mathbf{CA}_{out}^{c} &=  \mathbf{MHSA}_{output}^{c} + \mathbf{MHCA}_{output}^{c}
\end{aligned}
\end{equation}

\noindent Our proposed context consumption approach described in Fig.~\ref{fig:system} is very similar to hyper-prompting \cite{he2022hyperprompt}. However, unlike \cite{he2022hyperprompt}, we only have a single global prompt.~The number of global prompts can be extended towards various task specific fine-tuning use-cases and is left for future work. We concatenate $\mathbf{P}_{1:TW}$ to the acoustic representations $\mathbf{A}_{-CW:i}^{c}$ along the temporal axis. %
~Next we project the concatenated tensor with key and value kernels to obtain augmented keys $\hat{\mathbf{k}}^{c}$ and values $\hat{\mathbf{v}}^{c}$. These are used in a \gls{MHA} operation along with original queries $\mathbf{q}^{c}$ obtained from  $\mathbf{A}_{-CW:i}^{c}$ only. In this way we do not increase the effective acoustic temporal sequence length at the output of \gls{MHA}. We describe the entire operation in Equation \ref{eq:proposed_method}.

\begin{equation} \label{eq:proposed_method}
\begin{aligned}
\hat{\mathbf{k}}^{c} &= (\mathbf{P}_{1:TW} \oplus \mathbf{A}_{-CW:i}^{c} ) \mathbf{W}_{k}, \\
\hat{\mathbf{v}}^{c} &= (\mathbf{P}_{1:TW} \oplus \mathbf{A}_{-CW:i}^{c} ) \mathbf{W}_{v}, \\
\mathbf{q}^{c} &=  \mathbf{A}_{-CW:i}^{c} \mathbf{W}_{q}, \\
\mathbf{MHSA}_{output}^{c} &= \textbf{MHA}(\mathbf{q}^{c}, \hat{\mathbf{k}}^{c},\hat{\mathbf{v}}^{c})
\end{aligned}
\end{equation}

\noindent We believe our method has the following advantages over \gls{MHCA}: (i) The textual and acoustic kernels for query, key and value projections are shared, forcing the kernels to produce fused multimodal representations. (ii) The softmax is shared between textual context and regular acoustic context frames, thereby leading to better equalization of the final output representation of a particular acoustic frame.

\section{Experimental Setup}
\label{sec:experimetal_setup}

\subsection{Model configuration}
\label{ssec:model_configuration}
As a seed model for all experiments we use a \gls{RNN-T} conformer transducer with 91.7 M parameters and 12 conformer blocks.~The encoder dimension is 512 and each conformer block has 8 self-attention heads with head size 64. The encoder has in total 60.6 M parameters. We use $\mathbf{\gls{CW}=\mathbf{40}}$ for all our experiments.~The prediction network consist of 2 x 1280 \gls{LSTM} layers and it consists of 23.9 M parameters.~The joint network is a feed-forward layer with 512 units and tanh activation. A softmax layer is used on top with a word-piece vocabulary size of 4K.~The audio is processed into 64-dimensional Log-Mel-Frequency features per frame with frame shift of 10 ms downsampled
by a factor of 3. SpecAugment was used for feature-based augmentation during model training \cite{Park_2019}. The seed model was trained using Adam optimizer with a linear warm-up period (5K steps) and exponential decay thereafter \cite{DBLP:journals/corr/KingmaB14}.~The seed model was trained without context. We evaluate a baseline and our proposed methods for context modeling by fine-tuning the seed model without and with the presence of context \cite{hori,Alanis2022}.~Fine-tuning is conducted for 250K steps and the final model is obtained by averaging the last 10 checkpoints (5K steps each) \cite{DBLP:journals/corr/abs-2010-10504}.

\subsection{Datasets}
\label{ssec:datasets}
Models are trained with a de-identified in-house English voice assistant dataset consisting of audio and text transcriptions. A fine-tuning training set of 140K hours of audio with context is used.~Context is the recognition associated to the previous stream within a session. Sessions are defined by grouping streams from the same voice assistant device and using a sliding window of 90 seconds between two consecutive streams.~We estimated that up to 70\% of the utterances have non-empty context \gls{ASR} recognition.%
~We report \gls{rWERR} with respect to the baseline (fine-tuning without context). We show results on two in-house test sets: average traffic and multi-turn. The former contains 29.1K utterances, and 71\% of the utterances have non-empty context. The second dataset has 7K utterances with non-empty context for each utterance.~As reference, the fine-tuned baseline model improves by 11\% and 3.3\% \gls{rWERR} over the seed model in the multi-turn and average traffic test sets, respectively.~Performance of the seed model is below 10\% WER absolute.

\section{Experimental results}

\label{sec:experimetal_results}
\subsection{Context integration approaches}
\begin{table}[t]
\caption{\small Comparison of relative word error rate reduction (rWERR) achieved by different context consumption and context generation approaches.~\textit{CA} refers to Cross attention, \textit{CP} refers to \textit{copying} joint network params for first projection layer initialization.}
\label{results_bert_token}
    \centering
{\small %
   \begin{tabular}{|l|l|l|l|l|l|}
    \hline
       \footnotesize \textbf{Context cons.} &  \footnotesize   \textbf{Context gen.} &    \multirow{-2}{*}[-1ex]{\footnotesize \textbf{Multi-}} &
       \multirow{-2}{*}[-1ex]{\footnotesize \textbf{Avg.}} &
       \footnotesize  \textbf{Params} \\ 
       & & \footnotesize \textbf{Turn.} & \footnotesize \textbf{Traffic.} & \\ \hline
        None (Baseline) & None & 0.0\% & 0.0\% & +0.0\%    \\ \hline
        CA & BERT/Sent. & 3.2\% & 2.4\% & +15.1\% \\ \hline
        CA + CP & BERT/Tok. & 4.1\% & 3.3\% & +18.7\% \\ \hline
        Feature concat. & BERT/Sent. & 2.3\% & 1.4\% & +3.8\% \\ \hline 
          Prompts& BERT/Sent. & 3.4\% & 2.0\% & +0.5\% \\ \hline
         Prompts & BERT/Tok. & 5.0\% & 2.9\% & +0.5\% \\ \hline
        Prompts & SPM/Tok. & 4.6\% & 3.7\% & +3.7\% \\ \hline
        Prompts + CP & SPM/Tok. & 5.9\% & 3.4\% & +3.7\% \\ \hline
    \end{tabular}
}
\end{table}
\label{ssec:context_integration}
Table \ref{results_bert_token} compares the integration approaches explored in this paper. \gls{CP} refers to copying the joint network parameters in first projection layer as initialization.~All models with context provide significant gains in respect to the baseline.~Cross Attention (CA) and prompting variants using sentence embeddings outperform \gls{FC} by up to +1.8\% and +1.1\% \gls{rWERR} respectively.~This can be due to the lack of a mapping mechanism between textual and acoustic space in the \gls{FC} method.~A similar trend was observed in the average traffic test set.~All prompt token models consistently outperformed the equivalent \gls{CA} variant in the multi-turn test set, despite the significantly smaller increase in model parameters.~For instance, prompting using \gls{BERT} and \gls{SPM} tokens achieves 5.0\% and 5.9\% \gls{rWERR} respectively. The \gls{CA} model (with \gls{SPM} tokens) achieves 4.1\% \gls{rWERR}. This model adds +18.7\% parameters, while the prompting alternatives introduce only up to 3.7\% more. 
\begin{table}[!t]
\caption{Relative word error rate reduction (rWERR) on Average Traffic test set for selected models categorized by the presence or absence of context. All models use prompting for context consumption.}
    \centering
    {\small %
    \begin{tabular}{|l|l|l|l|l|l|}
    \hline
     \footnotesize   \textbf{Context gen.}  & \footnotesize \textbf{All} & \footnotesize \textbf{Context} & \footnotesize \textbf{No Context} \\ \hline
          None (Baseline) & 0.0\% & 0.0\% & 0.0\%  \\ \hline
        BERT/Sent.  & 2.0\% & 1.9\% & 0.9\%  \\ \hline
        BERT/Toks. & 2.9\% & 2.9\% & 1.6\%  \\ \hline
        SPM/Toks.  & 3.7\% & 4.1\% & 1.5\%  \\ \hline
        \textit{Dataset Size} & 29.1K & 20.7K & 8.4K \\ \hline
    \end{tabular}
    }
    \label{ablation_avg_context_vs_no_context}
\end{table}
\subsection{Textual Context Representation}
\label{ssec:textual_context_representation}
We observed token embeddings lead to better WER compared to sentence embeddings. For example, using \gls{BERT} tokens lead to an improvement of +1.6\% and +0.9\% \gls{rWERR} in the multi-turn and average traffic test sets over the sentence embedding counterpart. This indicates that the acoustic encoder benefits from the finer level of granularity provided by distinct sub-words representations. Similar trends have been observed for textual encoders \cite{li-etal-2020-sentence}. We also found \gls{BERT} marginally outperforms \gls{SPM} embeddings (+0.4\% \gls{rWERR} in the multi-turn test set) despite the smaller amount of added parameters to the model (+0.5\% vs +3.7\%). This may be the result of the \gls{BERT} model having seen more textual data and providing better textual representations than the trained \gls{SPM} projection layers. On the other hand, results were comparable in the average traffic test set. The best results were obtained with the SPM encoder when the first projection layer is initialized with the weights of the prediction network embedding layer (i.e. \textit{copying}). This initialization enables encoding text with the same word piece vocabulary of the base model, avoiding a vocabulary mismatch. This approach is advantageous because it reduces the need to use an external encoder. Despite the slight increase in encoder parameters compared to the \gls{BERT} variant, the total number of OPs is smaller since we do not need to run a \gls{BERT} forward pass when encoding the text, reducing overall inference latency.

Table \ref{ablation_avg_context_vs_no_context} shows \gls{rWERR} results for a breakdown of the average traffic test set based on whether there is or not context available. Values reported are in respect to the \gls{WER} of the baseline on each subset.~We observed that the majority of the gains are obtained on the subset of utterances having context available. For instance, the model trained with the \gls{SPM} encoder improves over the baseline by 3.1\% \gls{rWERR} on the subset of utterances having context. Candidates also improve up to 1.6\% on the subset of utterances without context, which shows our approach is robust for the cases when context is not available.

\begin{table}[!t]
\caption{Relative word error rate reduction (rWERR) when fine-tuning different model components. We utilized the model using SPM embeddings and copying.~\textit{All} refers to fine-tuning the full conformer-transducer model. Fine tuning was carried out for 250K steps unless a different value is specified. WERs are relative to fine-tuning without context.}
    \centering
    {\small %
    \begin{tabular}{|l|l|l|l|}
    \hline
       \footnotesize \textbf{Model setup} &  \footnotesize\textbf{Multi-turn} &  \footnotesize \textbf{Avg. Traffic} & \footnotesize \textbf{Train. params.}  \\ \hline
        Baseline  & 0.0\% & 0.0\% & 91.7M  \\ \hline
        All & 5.9\% & 3.4\% & 103.5M \\ \hline
        MHAs and projs.  & 5.8\% & 3.1\% & 11.7M (11.4\%)  \\ \hline
        Only projs. & 3.5\% & 2.5\% & 2.3M (2.2\%)  \\ \hline
        Only projs. (25K) & 3.7\% & 2.3\% & 2.3M (2.2\%)  \\ \hline
        Only projs. (50K) & 3.9\% & 2.3\% & 2.3M (2.2\%)  \\ \hline
    \end{tabular}
    }
    \label{ablation_training}
\end{table}

\subsection{Training regime}
\label{ssec:training_regime}
Table \ref{ablation_training} provides a comparison when not all components of the \gls{RNN-T} model are fine-tuned. Results are shown for the model trained with SPM encoder with \textit{copying}. We found that by fine-tuning only the \gls{MHA} modules and projection layers, almost the same performance can be achieved, with only negligible changes of -0.1\% and -0.4\% \gls{rWERR} (in respect to fine-tuning all parameters) in the multi-turn and average traffic test set.~The larger gap in the latter test set can be explained by the fact that fine-tuning other \gls{RNN-T} modules contributes to overall performance gains. This result also suggests that adapting the attention mechanism is sufficient for the model to incorporate context effectively. This training strategy provides faster training speed given only 11.4\% of the model parameters are fine-tuned.
We also experimented with fine-tuning only the projection layers.~We found significant gains are still obtained in the multi-turn test set, when fine-tuning for 50k steps (3.9\% \gls{WERR}) and for 25K steps (3.5\% \gls{WERR}).~However,~there was a loss of performance of -1.9\% \gls{WERR} in respect to the best performing model in the multi-turn test set.~We noticed training starts to overfit after 50K.~This result demonstrates our approach is also effective incorporating context for the scenario when modifying the base model parameters is not feasible.~Under this regime the training time is reduced significantly since only up to 2.2M parameters need to be trained for a limited number of steps.

\section{Conclusions}
\label{sec:conclusions}
In this work, we proposed to integrate context in the \gls{RNN-T} model by prompting textual representations to obtain fused multimodal representations.~We demonstrated this approach consistently outperforms cross-attention and feature concatenation,~while being more cost-effective.~We proposed a simple, yet effective, mechanism that re-uses the internal \gls{RNN-T} embedding projection network to create text representations.~This method outperforms representations obtained with a \gls{BERT} model, and has the additional benefit of incurring in less ops.~Our best candidate achieves 5.9\% \gls{rWERR} in a multi-turn test set,~without incurring degradations on non-contextual use cases.~We also showed our method can achieve comparable results when only the \gls{MHSA} and projections are fined-tuned.

\bibliographystyle{abbrv}
\bibliography{strings,refs}

\end{document}